\UseRawInputEncoding
\documentclass[letterpaper, 10 pt, conference]{ieeeconf}  
\usepackage[ruled]{algorithm2e} 
\usepackage{subfigure}
\usepackage{array} 
\usepackage{makecell}
\usepackage{amsmath}
\usepackage{pifont}
\usepackage{setspace}
\usepackage{xcolor}
\usepackage{units}
\usepackage{amsfonts}

\IEEEoverridecommandlockouts                              

\usepackage{graphics} 
\usepackage{graphicx}
\usepackage{color}
\usepackage[colorlinks,linkcolor=blue,anchorcolor=blue,citecolor=blue]{hyperref}

\title{\LARGE \bf
	Task2Morph: Differentiable Task-inspired Framework  \\ for Contact-Aware Robot Design
}

\author{Yishuai Cai$^{\dag}$, Shaowu Yang$^{\dag}$, Minglong Li$^{*}$, Xinglin Chen, Yunxin Mao, Xiaodong Yi and Wenjing Yang  
	\thanks{$^{\dag}$Equal contribution.}
	\thanks{*Corresponding author.}
	\thanks{All the authors are with the Institute for Quantum Information State Key Laboratory of High Performance Computing, College of Computer Science and Technology, National University of Defense Technology, Changsha, China., {\tt\small \{caiyishuai, shaowu.yang, liminglong10, chenxinglin, maoyunxin, wenjing.yang\}@nudt.edu.cn, xdong\_yi@163.com}}%
}

\begin{document}

	\maketitle
	\thispagestyle{empty}
	\pagestyle{empty}
	
	
	
 	\begin{abstract}
		Optimizing the morphologies and the controllers that adapt to various tasks is a critical issue in the field of robot design, aka. embodied intelligence. Previous works typically model it as a joint optimization problem and use search-based methods to find the optimal solution in the morphology space. However, they ignore the implicit knowledge of task-to-morphology mapping which can directly inspire robot design. For example, flipping heavier boxes tends to require more muscular robot arms. This paper proposes a novel and general differentiable task-inspired framework for contact-aware robot design called Task2Morph. We abstract task features highly related to task performance and use them to build a task-to-morphology mapping. Further, we embed the mapping into a differentiable robot design process, where the gradient information is leveraged for both the mapping learning and the whole optimization. The experiments are conducted on three scenarios, and the results validate that Task2Morph outperforms DiffHand, which lacks a task-inspired morphology module, in terms of efficiency and effectiveness.
		
		
		%
		
		
		
	\end{abstract}

	\section{INTRODUCTION}
	
	Intelligent behaviors tend to be learned more rapidly by agents whose morphologies are better adapted to their environment \cite{bongard2014morphology,hejna2021task,gupta2022metamorph}. The physical manifestation of a robot plays an important role in shaping its behavioral and cognitive abilities, aka. embodied intelligence \cite{gupta2021embodied,nygaard2021real,wilson2013embodied,kacprzyk2015springer}. However, the artificial intelligence (AI) community has mainly focused on disembodied cognition, such as computer vision, natural language processing, and video games, etc. \cite{yu2021interaction, gupta2021embodied}, and has paid less attention to embodied intelligence. Designing artificial embodied agents, i.e., robot design, with well-adapted morphologies that can learn control tasks in diverse complex tasks is challenging due to: \emph{the problem complexity of embodied intelligence scales across three axes, including task, morphology, and controller} \cite{gupta2021embodied}.
	
	To solve that, previous studies \cite{wang2019neural,zhao2020robogrammar,gupta2021embodied} typically model it as a joint optimization problem with an interactive two-layer loop, that is, the outer layer optimizes the morphology while the inner layer optimizes the control. They usually begin with a fixed or random morphology for any new task, and repeatedly search and evaluate in the morphology design space to find the optimal one \cite{whitman2020nips}. However, they ignore the latent mapping relationship between tasks and optimal morphologies, which leads to exponential complexity. In other words, given a well-trained task-to-morphology mapping, the optimal morphology corresponding to the task can be obtained directly. As a result, the two-layer loop can be merged into one, which can greatly reduce the problem complexity.
	
	\begin{figure}[t]
		\centering
		\includegraphics[scale=0.17]{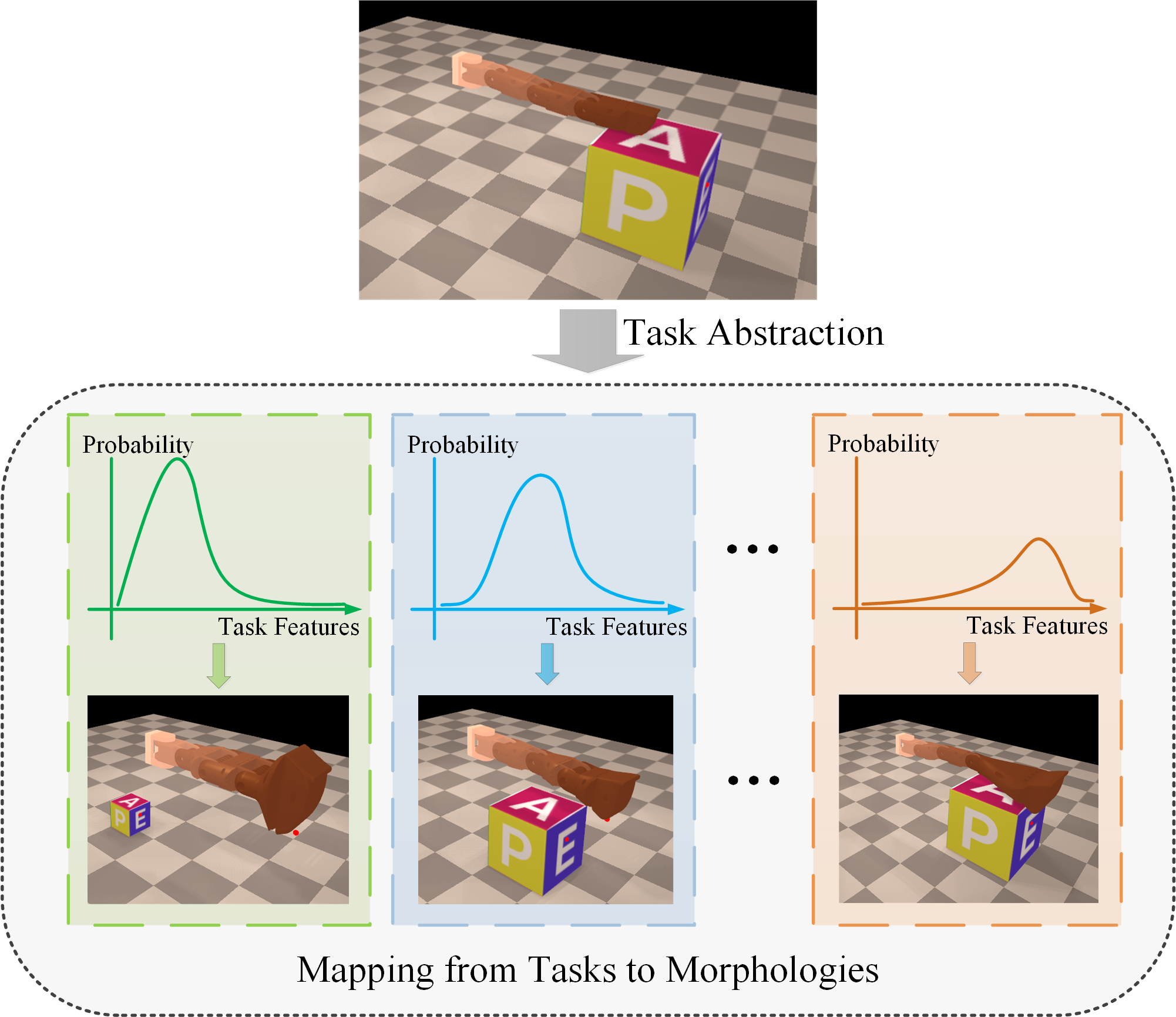}
		\caption{In the scenario of box flipping, the position and size of the box can be regarded as task features since they are highly correlated with task performance. For different boxes, the feature distribution is different, which affects the optimal morphology of the robot. We aim to abstract the related task features and use them to inspire robot design directly.}
		\label{describtion}
	\end{figure}

	
	To further illustrate the above argument, consider a box-flipping scenario where a robot arm's morphology is to be designed to flip boxes more efficiently, as shown in Fig. \ref{describtion}. Previous research has maintained the initial design configuration unchanged when the task features (including object features and task objectives, etc.) change. This can lead to lengthy optimization times or being trapped in local optima. While many studies such as \cite{schaff2019jointly, yuantransform2act} focus on exploring better optimization methods, there is a significant correlation between task features and the resulting morphology, as the robot continually interacts with the environment and achieves the task as the optimization objective. This is particularly evident in contact-based tasks like box-flipping, where the robot directly interacts with the task objects. Thus, we propose using task features to initialize the morphology parameters, namely task-inspired morphology, such as longer limbs for distant boxes or more muscular limbs for larger boxes. 

	Furthermore, we do not need to consider every detail in the box-flipping tasks, such as box color. Instead, we should focus on task features highly correlated with task performance, such as box size and position. In essence, the task implicitly provides critical knowledge that can inform optimal morphology design. Therefore, we introduce a practical approach: \emph{abstract performance-related task features and use them to model a task-to-morphology mapping}.
	
	However, there remains another challenge that how to optimize the entire process, including the mapping. Many works use data-inefficient gradient-free methods\cite{xu2021end}, such as graph search or genetic algorithm, to optimize morphology \cite{gupta2021embodied,zhao2020robogrammar,wang2019neural,schaff2019jointly}, where experience is not shared between agents in each iteration \cite{yuantransform2act}. If a differentiable joint optimization method is available, then the morphology optimization can be more efficient and accurate since it will have an easy-to-get direction, i.e., gradient information. More importantly, the learning of task-to-morphology mapping can be embedded into the optimization process and benefit from gradient backpropagation.

    Fortunately, there is also a lot of work \cite{spielberg2017functional,bravo2020one,ha2017joint,chen2021hardware} to optimize morphology by leveraging analytic gradients. However, in some studies \cite{yuantransform2act}, morphology is simplified as a composition of primitive shapes (i.e., cylinders, cuboids), and in some cases \cite{ha2017joint,chen2021hardware}, only the parameters of simple component properties (such as link length, spring stiffness, etc.) are optimized. These approaches may not ensure a diverse morphology space, which is crucial for contact-aware tasks that demand higher geometric requirements on morphology. Among them, Xu et al. \cite{xu2021end} propose an end-to-end differentiable morphology design framework called \emph{DiffHand}. In this framework, they introduce a continuous morphology representation based on Cage-Based Deformation (CBD), which works for a wide range of complex shapes. They also develop a differentiable simulator to make the full robot design framework differentiable. Thus, the gradient information can be leveraged as the direction to guide the optimization.
	
	
	
	This paper proposes \emph{Task2Morph}, a general differentiable task-inspired framework for contact-aware robot design. We introduce the task abstraction and task-to-morphology mapping into robot design and formalize the mapping as a regression problem between tasks and robot morphologies. Further, we embed the task-to-morphology mapping into \emph{DiffHand} to construct a novel and effective differentiable robot design method. In addition, our approach enables learning the mapping between various task features and their corresponding optimal morphologies while optimizing both the robot morphology and controller. Furthermore, it is inherently more suitable for morphology adaptation scenarios in which task features may vary, such as sorting through multiple boxes of different sizes and positions. In such scenarios, a good initial morphology can be generated for each distinct task features (e.g., box size or position) based on the learned task-to-morphology mapping, and subsequently fine-tuned to obtain the optimal final morphology.
	
	Note that \emph{Task2Morph} differs from \emph{DiffHand}. On the one hand, \emph{Task2Morph} introduces the idea of task-inspired morphology into the algorithm implementation, focusing on task abstraction and the construction of mapping between task and morphology. On the other hand, \emph{Task2Morph} integrates mapping optimization into the gradient backpropagation process to ensure the new framework remains fully differentiable end-to-end.
	
	
	
	Our contributions are summarized as follows:
	
	\begin{itemize}
		
		\item We introduce task abstraction and incorporate a task-to-morphology mapping into the robot design process, which is particularly suitable for morphology adaptation tasks with variable features.
		
		\item We propose a general differentiable task-inspired robot design framework, where the gradient information is used to optimize both the task-to-morphology mapping and the whole robot design process.
		
		\item We conduct the experiments on three scenarios inspired by Xu et al. \cite{xu2021end}. The results validate that \emph{Task2Morph} outperforms \emph{DiffHand} in efficiency and efficacy.

	\end{itemize}

	\begin{figure*}[thpb]
	\centering
	\includegraphics[scale=0.287]{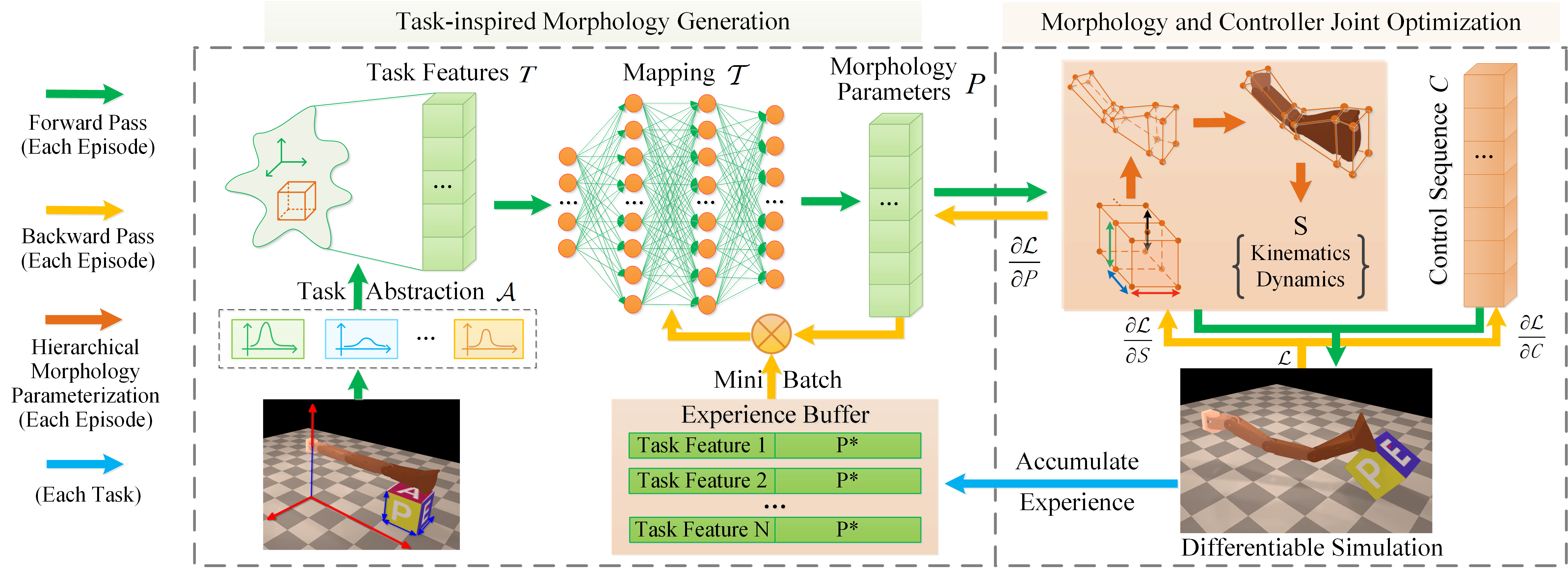}
	\caption{\textbf{Task2Morph Overview.} It is mainly composed of two parts, the left part is task-inspired morphology generation which is the focus of our work. It consists of three parts: task abstraction, mapping from task features to morphology parameters and experiment backpropagation. The right part showcases the co-design framework known as \emph{DiffHand}, which is used to fine-tune our initial morphology and optimize the controller.}
	\label{struct}
	\end{figure*}

	\section{RELATED WORK}
	
	 
	
	Thanks to the fast-growing machine learning and intelligent control, various morphology automation design methods can generate robot morphologies with a high level of adaptability for specific tasks and environments.\cite{nygaard2021real,schaff2019jointly}. 
	
	\textbf{Co-optimization of Robot Design and Control}.
	Optimizing robot design is typically formulated as a bi-level problem, with the outer level optimizing morphology and the inner level optimizing the controller \cite{yuantransform2act,chen2022c}. Evolutionary strategies, such as those proposed by Sims \cite{sims1994evolving} and Gupta et al. \cite{gupta2021embodied}, are commonly used for these optimization algorithms. Wang et al. \cite{wang2019neural} introduced neural graph evolution, which uses simple mutational primitives to evolve graph structures. Zhao et al. \cite{zhao2020robogrammar} proposed RoboGrammer, which uses graph heuristic search to determine optimal robot morphology for traversing various terrains.
	Bhatia et al. \cite{bhatia2021evolution} developed Evolution Gym, a benchmark for morphology and controller co-design of soft robots, which uses CPPN-NEAT to evolve robot morphology.

	Although the topologies of the above work are diverse, the morphology of the limbs is oversimplified and modeled as a combination of simple primitive shapes (e.g., cylinders, cuboids, etc.) \cite{xu2021end}. In addition, while evolutionary computation is promising, it has lower sample efficiency and higher computational complexity compared to gradient-based optimization methods \cite{xu2021end,chen2021hardware,vemula2019contrasting}. This is primarily attributed to the absence of information sharing among individuals in the population, resulting in significant data wastage during the elimination of poor individuals throughout the iteration process \cite{yuantransform2act}.
	
	
	%
	
	\textbf{Gradient-based Methods}.
	In light of this, there are many studies \cite{spielberg2017functional,ha2017joint} that adopt gradient-based methods to optimize morphology to improve sample efficiency. Chen et al. \cite{chen2021hardware} formulate the mechanical design as a set of parameters that can be optimized using gradients, and jointly optimize them with the control policy using deep reinforcement learning. Yuan et al. \cite{yuantransform2act} regard the generation of robot body components as one available action in the decision-making process, and adopt reinforcement learning based on strategy gradient to solve the problem. However, their morphological representation still has the problem of oversimplification (only involving the length and thickness of components), and cannot represent rich morphological space, especially for contact tasks with high geometric shapes. Fortunately, Xu et al. \cite{xu2021end} proposed \emph{DiffHand}, which includes a CBD shape representation and a differentiable rigid-body simulator. This shape representation ensures rich geometric shapes of components in fixed topologies. More importantly, their simulator provides analytic gradients for shape optimization, which significantly improves the optimization efficiency compared to existing gradient-free methods.
	



	
	
	Additionally, prior studies on morphology adaptation, including \emph{DiffHand}, required starting from a fixed or randomly initialized morphology and optimizing it separately for each new task, even if only the task features changed, which incurred significant computational cost for each search \cite{whitman2020nips}. They ignore the implicit knowledge in the task-to-morphology mapping that could directly inspire robot design.

	\textbf{Task-inspired Morphology Generation}.
	There are a few studies on task abstraction and task-inspired morphology generation. 
	Hejna III et al. \cite{hejna2021task} employ an information-theoretic objective to evaluate potential initial morphology but do not consider task features. Schaff and Walter \cite{schaff2022n} propose the co-optimizing method N-LIMB, which involves sparse terrain coding to optimize control strategies rather than morphology. Nygaard et al. \cite{nygaard2021real} develop a quadruped robot that modifies its discrete morphology combinations based on the hardness and roughness of the ground as task inputs. Whitman et al. \cite{whitman2020nips,whitman2020aaai} and Hu et al. \cite{hu2022modular} use neural networks to map two-dimensional heat maps of various terrains onto modular robots. While modular robots are promising due to their module reusability, their discrete module combination space may not be sufficient for contact tasks that require more nuanced and flexible geometric optimization. Additionally, compared to Ha et al. \cite{ha2021fit2form}, who utilize a data-driven approach to construct an end-to-end mapping of object to 3D gripper shape design, our approach covers a broader range of contact tasks and incorporates joint optimization of both morphology and controllers. Moreover, our approach employs manual knowledge to directly extract and encode performance-related task features(e.g., the position/size of objects), reducing the distortion of feature extraction and preserving interpretability.
	
	
	
	Our approach, \emph{Task2Morph}, is based on \emph{DiffHand} and specifically designed for contact-aware tasks. It covers a rich geometric space while learning continuous morphology design for different task features through gradient-based methods, thereby extracting knowledge that is beneficial for morphology optimization.

\begin{algorithm}[t]\small
	\label{Task2Morph}
	\caption{Task2Morph} 
	\LinesNumbered  
	\KwIn{A task} 
	\KwOut{The best morphology parameters $P^{*}$ and control sequence $C^{*}$}
	
	Initialize the regression model $\mathcal{T} _{\theta }(\cdot) $ with random parameters $\theta$\; 
	Initialize control sequence $C$\ with a random vector;\\
	\textcolor{purple}{$\triangleright$ Generate task-inspired morphology }\\
	\begin{spacing}{1}
		$T\leftarrow \mathcal{A}(task) $\; 
		$P\leftarrow \mathcal{T} _{\theta }(T) $\;
	\end{spacing}
	\textcolor{purple}{$\triangleright$ Fine-tuning morphology}\\
	\For{$i\leftarrow1$ \textbf{to} $iter_{max}$}{
		\begin{spacing}{1}
			$S\leftarrow \mathcal{F}(P)$\;
		\end{spacing}
		\begin{spacing}{1.25}
			$\mathcal{L},
			\frac{\partial{\mathcal{L}}}{\partial S}, 
			\frac{\partial{\mathcal{L}}}{\partial C} \leftarrow Simulation(S,C)$\;
		\end{spacing}
		\If{convergence
			$(\frac{\partial{\mathcal{L}}}{\partial S} ,
			\frac{\partial{\mathcal{L}}}{\partial C})$
		}{
			break\;
		}
		\begin{spacing}{1.25}
			$\frac{\partial{\mathcal{L}}}{\partial P} 
			\leftarrow \frac{\partial{S}}{\partial P}
			\frac{\partial{\mathcal{L}}}{\partial S}$\;
			
			$C\leftarrow\emph{L-BFGS-B}(C,\mathcal{L},\frac{\partial{\mathcal{L}}}{\partial C})$\;
		\end{spacing}
		$P\leftarrow\emph{L-BFGS-B}(P,\mathcal{L},\frac{\partial{\mathcal{L}}}{\partial P})$\;
		
		\textcolor{purple}{$\triangleright$ Training regression model}\\
		Sample a mini-batch $D$ of historical task $T_{k}$ and corresponding optimal $P_{k}^{*}$ of size $K$\;
		\begin{spacing}{1.1}
			Update $\mathcal{T} _{\theta }(\cdot) $ one step by \emph{SGD} with loss:\
		\end{spacing}
		\small{$\alpha{\begin{Vmatrix}
					P-\mathcal{T} _{\theta }(T)
			\end{Vmatrix}}^{2}+
			(1-\alpha)\sum_{T_k,P_k^*\in D}{\begin{Vmatrix}
					P_k^*-\mathcal{T} _{\theta }(T_k)
			\end{Vmatrix}}^{2}$}
	}
	$P^*\leftarrow P$, $C^*\leftarrow C$\;
	\If{$\mathcal{L} < \mathcal{L}_{max}$}{
		Put $T,P^*$ into \emph{buffer}
	}	
\end{algorithm}

	\section{METHOD}
	Here, we describe our differentiable task-inspired framework \emph{Task2Morph} for contact-aware robot design. It directly generates a more suitable morphology by learning the task-to-morphology mapping during the joint optimization of robot morphology and controller, hence enhancing its future adaptability to new tasks. More specifically, as shown in Figure \ref{struct}, it consists of two modules: task-inspired morphology generation and morphology and controller joint optimization. Our work focuses on the former, which includes task abstraction, mapping from tasks to morphologies and experience backpropagation. (see Section III-B and Section III-C for more details). In the latter, we adopt the co-design framework \emph{DiffHand} \cite{xu2021end} to fine-tune and evaluate our initial morphology (see Section III-A for more details). Algorithm \ref{Task2Morph} gives an outline of \emph{Task2Morph} implementation.
	
	
	In this paper, the term ``different tasks" refers to tasks within the same scenario that share a common objective but exhibit varying object features. For example, in the box-flipping scenario, we consider it a different task when the size and location of the box differ, even if the goal remains the same.

	\subsection{DiffHand}
	This section briefly introduces two algorithms in \emph{DiffHand} \cite{xu2021end}: continuous morphology representation based on CBD and a differentiable simulator.	
	
	\textbf{Morphology Representation}. 
	CBD achieves real-time deformation of internal meshes by controlling the positions of the outer cage vertices (handles). It uses hierarchical morphology parameterization\cite{xu2021end} to represent a variety of complex robot configurations with fewer continuous shape parameters.
	In Figure \ref{struct}, the high-level shape parameters are defined by the positions of the cage handles, denoted as $P$. These handles include both the cage vertices and the center points of each face. For example, a cube-shaped cage has 14 cage handles (8 vertices and 6 faces). Adjusting the positions of the cage handles (indicated by red, green, blue, and black arrows) , enables deformation and scaling of the internal meshes.  These meshes constitute the low-level shape parameters, represented by $M$. The initial meshes are obtained from a component database designed by Xu et al. \cite{xu2021end}, which contains meshes of various components and their corresponding preset cages.
	
	When the high-level shape parameters change, the low-level shape parameters also change while preserving their geometric features. The average value coordinate method [31] is employed in CBD to compute a normalized barycentric coordinate $\omega \in R^{|P|}$, known as deformation weights, for any arbitrary point $\tilde{m} \in M$. This is expressed by the equation:
	\begin{align}
	\label{eq1}
	\tilde{m}=\sum_j^{|P|} \boldsymbol{w}_j \tilde{P}(j) \quad and \quad \sum_j^{|P|} \boldsymbol{w}_j=1
	\end{align}
	Here, $\tilde{P}(j)$ represents the position of the $j$-th cage handle in $P$. The computed deformation weights can be reused. Given the known positions of high-level cage handles, the new meshes (low-level shape parameters) can be obtained by computing:
	\begin{align}
	\label{eq2}
	m=\sum_j^{|P|} \boldsymbol{w}_j P(j) \quad
	\end{align}
	where $P(j)$ represents the new positions of the cage handles, and $m$ is the new position of $\tilde{m}$ under deformation. This linear combination enables a fully differentiable mapping from high-level shape parameters to low-level shape parameters.
	When using the low-level mesh parameters $M$ for simulation, they need to be transformed into the corresponding low-level kinematics and dynamics parameters of the robot, denoted as $S$, through an analytical function. $S$ includes kinematic and dynamic parameters, and the implementation of the analytical function uses deformation parameterization techniques and other kinematic and dynamic computation methods.
	Importantly, the entire process of morphology parameterization, from the initial high-level cage parameters to simulation parameters, is differentiable. This enables the gradient to be passed through the chain rule. Mathematically, the morphology parameter transformation from $P $ to $S $ is defined as $\mathcal{F}
	\colon \mathbb{R}^{|P|} \to \mathbb{R}^{|S|}$.
	\begin{align}
	\label{eq3}
	S =  \mathcal{F}(P) 
	\end{align}\indent 
	\textbf{Differentiable Simulator}.
	The simulator, based on RedMax \cite{wang2019redmax}, compactly expresses the equations of motion using reduced coordinates with minimal degrees of freedom. It integrates the BDF2 scheme for implicit time integration, utilizing SDIRK2 for the initial step. \cite{wang2019redmax}. By inputting simulation parameters $S$ and control sequences $C$, robots can perform tasks and evaluate efficiency using the task-specific objective loss $\mathcal{L}$. The adjoint sensitivity method \cite{geilinger2020add} is employed to compute simulation derivatives $\frac{\partial \mathcal{L}}{\partial S}$ and $\frac{\partial \mathcal{L}}{\partial C}$, allowing for backpropagation of the loss gradient through the entire pipeline, including morphology and control parameters. This encompasses the high-level cage parameters $P$ in the initial layer, which can be computed using the equation:
	\begin{align}
		\label{eq4}
		\frac{\partial \mathcal{L}}{\partial P} = \frac{\partial \mathcal{L}}{\partial S} \frac{\partial S}{\partial P}
	\end{align}
	For more detailed information, please refer to Xu et al. \cite{xu2021end}.

	\subsection{Task-inspired Morphology}
	\textbf{Task Abstraction.}
	The mapping relationship between tasks and corresponding optimal morphologies is challenging to find because of the vast task space and morphology space. We decide to start with task abstraction and extract features related to task completion effects. These features are typically obtained through perception algorithms \cite{wang2021joint} from computer vision or other related fields. However, in certain scenarios, robots can directly receive task-specific information, such as inputting the component size and punching position to an industrial robot to improve the accuracy of component punching. Therefore, in this paper, instead of relying on perception algorithms, we adopt a manual knowledge-based approach to extract task features. We enumerate the object features (such as shape, color, position, etc.) involved in the task and identify those that may influence the robot's ability to accomplish the task as performance-related task features. These features are denoted as $ T $ and the process of task abstraction is represented by $\mathcal{A} $. Moreover, we employ high-level cage parameters $P$ as our morphology parameters since it can represent a huge morphology space with fewer parameters.
	
	 

	
	\textbf{Mapping from Tasks to Morphologies.}
	As the task features and morphology parameters are continuous, we formalize the problem of finding the mapping between the task features and the optimal morphology parameters. We define the mapping of task feature $ T $ to the corresponding optimal morphology parameter $P^*$ as $\mathcal{T}^*:\mathbb{R}^{|T|}\to \mathbb{R}^{|P|}, i.e.,P^* =\mathcal{T}^{*}(T)$. We expect the regression model $\mathcal{T}$ to be as close as possible to a mapping $ \mathcal{T}^* $. Deep learning has shown outstanding ability in learning problems and is particularly suited for solving multiple regression problems. Therefore, we use neural networks (NN) as regression models $\mathcal{T}$ with task features as inputs and optimal morphology parameters as outputs. Mathematically speaking, $ \mathcal{T}:\mathbb{R}^{|T|}\to \mathbb{R}^{|P|}$, 
	\begin{align}
	\label{eq5}
	P =  \mathcal{T}(T) 
	\end{align}\indent 
	\textbf{Morphology Fine-tuning.}
	When the model is not fully trained, the morphology parameters $ P $ output by the regression model are only used as the initial parameters since it is not always the optimal parameter $P^{*}$. Therefore, fine-tuning morphology is necessary at this stage. We use \emph{DiffHand} to implement morphology fine-tuning, taking $ P $ as the parameter at the beginning of the iteration and initializing the controller parameters $ C $. With the help of L-BFGS-B method \cite{xu2021end}, $ P $ and $ C $ are jointly optimized by computing the gradients $ \frac{\partial \mathcal{L}}{\partial P} $ and $ \frac{\partial \mathcal{L}}{\partial C} $, as illustrated in Algorithm \ref{Task2Morph}. After the algorithm converges, the final optimal morphology parameters are obtained. It is worth noting that morphology fine-tuning is a gradual procedure that requires multiple iterations. Nevertheless, the initial morphology parameters generated by a well-trained regression model typically require minimal or even no adjustments.
	

	\subsection{Experience Backpropagation}
	In addition to designing suitable morphologies for known tasks, our objective is to enable the regression model to handle new situations based on past data. Our learning-based approach relies on the backpropagation of gradients, which is critical in our framework, ensuring that tasks inspire morphologies. As demonstrated in Section III-B, the task is abstracted and parameterized in Equation \ref{eq5}, and the gradient transfer is presented in Equation \ref{eq4}. We incorporate the gradient $\frac{\partial \mathcal{L}}{\partial P}$ into the regression model by defining its loss function. In this paper, we adopt online learning to update the mapping, enabling robots to update themselves in real-time while performing tasks. During each iteration of the task, the differentiable simulator returns the gradients of the loss function with respect to the morphology and control sequence for the current task. We denote this part of the loss function as $L_{current}$, which is defined as: 
	\begin{align}
	\label{eq6}
	L_{current} = \parallel P_{current}-\mathcal{T}(T_{current}) \parallel^2
	\end{align}\indent 
	If only the gradient of the current loss with respect to the current morphology is considered, the network parameters may oscillate during the update process, leading to incomplete training of the model. Therefore, we introduce \emph{experience buffer} to store the past computed task features $T_k$ and their corresponding optimal morphology parameters$P_k^*$. This approach is similar to the \emph{experience replay buffer} \cite{tom2015replay} used in reinforcement learning, where a portion of the data is sampled from the \emph{experience buffer} at each episode to cooperatively update the regression model. For the portion of sampled data, its loss $L_{minibatch}$ is as follows:
	\begin{align}
	\label{eq7}
	L_{minibatch} = \sum_{{T_k,P_k^*}\in D}^{} \parallel P_k^*-\mathcal{T}(T_k) \parallel^2
	\end{align}
	where $D$ denotes a mini-batch of historical data sampled from the \emph{experience buffer}, $ \mathcal{T}(T_k) $ represents the output when inputting $T_k$ into the regression model. The loss function of the regression model consists of the aforementioned loss components (Equation \ref{eq6} and Equation \ref{eq7}), where the weight coefficient is denoted by $\alpha$. And the total loss is defined as:
	\begin{align}
	\label{eq8}
	L_{total} = \alpha L_{current}+ (1-\alpha) L_{minibatch}
	\end{align}

	\begin{table*}[h] 
	\vspace{0.2cm}
	\setlength{\abovecaptionskip}{0cm} 
	\setlength{\belowcaptionskip}{-0.2cm}
	\caption{Task Objectives and Task Features of Three Scenarios}
	\resizebox{\linewidth}{!}{
		\renewcommand\arraystretch{2}
			\begin{tabular}{|c|c|c|c|c|}
				\hline
				Scenarios & Objectives & Task Features & \makecell[c]{Number of \\ Task Features} & Range of Features\\ 
				\hline
				Finger Reach & \makecell[c]{Finger fixed to the wall touches four \\  designated spots in the air in sequence}  & \makecell[c]{Three-dimensional coordinates of the  four \\ target points $(x_i,y_i,z_i),i\in[0,3]$}& 12 & $x_i\in[-5,25], \quad y_i,z_i\in[-25,25]$ \\
				\hline
				Flip Box & \makecell[c]{Finger fixed at a certain height turns the \\ designated box 90$^\circ$ with as little effort as possible}  & \makecell[c]{Two-dimensional position coordinates and length, \\width and height of the box  $(x_{box},y_{box},a,b,c)$ }  & 5 & \makecell[c]{$x_{box}\in[-25,25],\quad y_{box}\in[-5,5],$ \\ $\quad a,b,c\in[2,9]$} \\
				\hline
				Rotate Plank & \makecell[c]{Fixed but moveable finger \\ rotate the plank on the stake 90$^\circ$} & \makecell[c]{Two-dimensional position coordinates of \\ the stake below the plank  $(x_{stake},y_{stake})$} & 2 & $x_{stake}\in[-5,25],  \quad y_{stake}\in[-10,10]$ \\
				\hline
			\end{tabular}
		}
	\label{objectives}
\end{table*}
	
	\begin{figure*}[t]
		\centering
		\subfigure
		{
			\begin{minipage}[b]{.31\linewidth}
				\centering
				\includegraphics[scale=0.36]{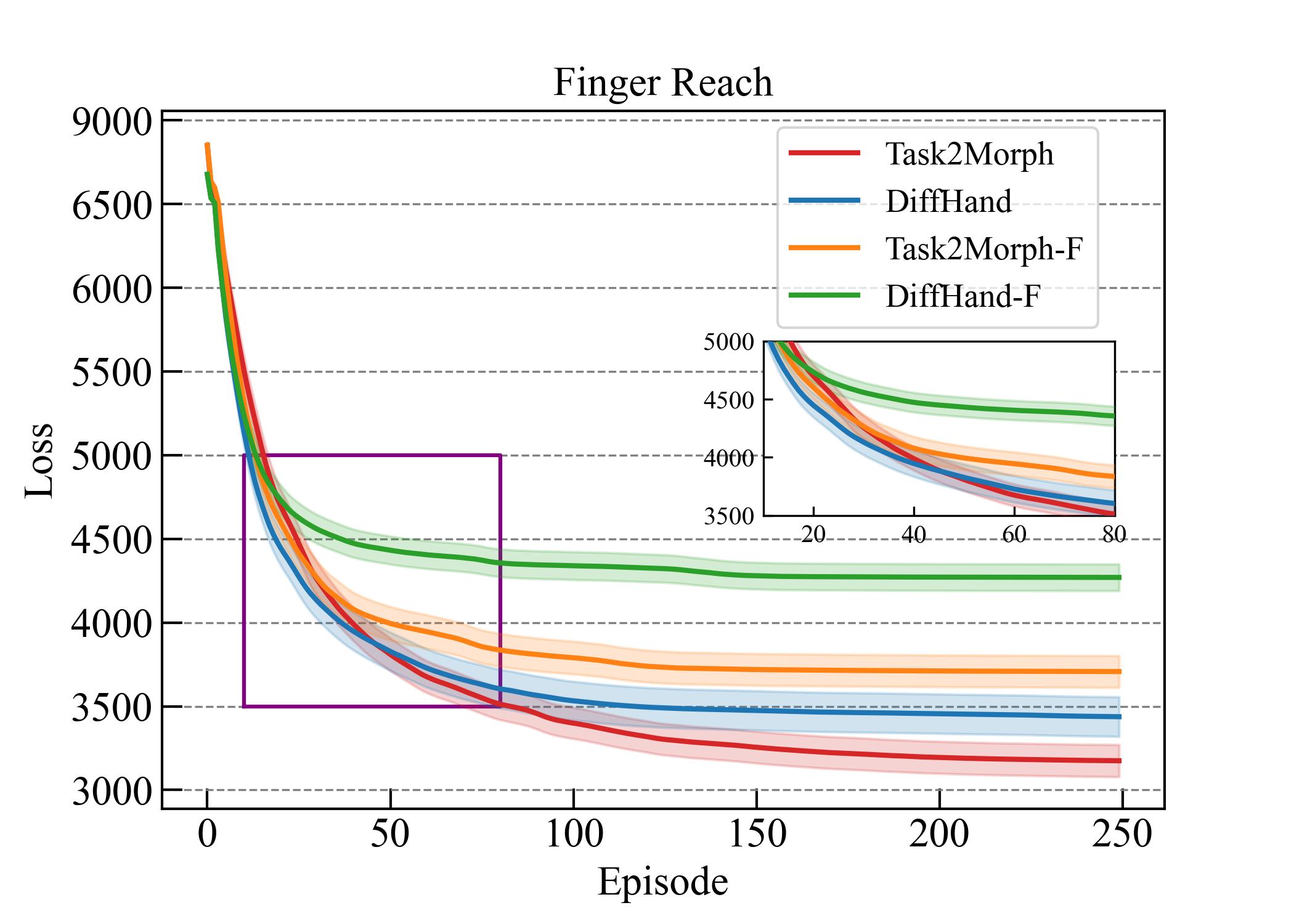} \\
			\end{minipage}
		}
		\subfigure
		{
			\begin{minipage}[b]{.31\linewidth}
				\centering
				\includegraphics[scale=0.36]{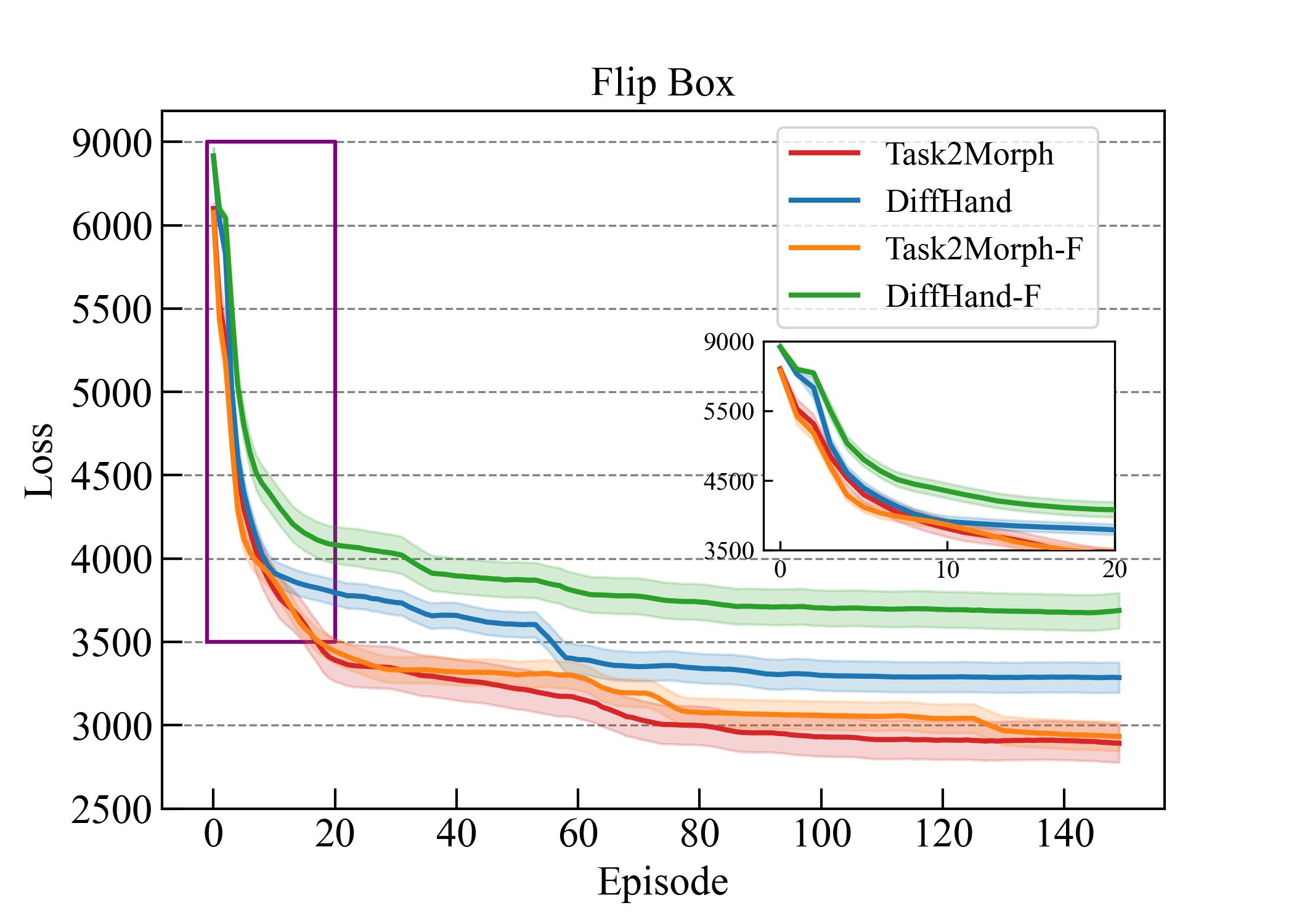} \\
			\end{minipage}
		}
		\subfigure
		{
			\begin{minipage}[b]{.31\linewidth}
				\centering
				\includegraphics[scale=0.36]{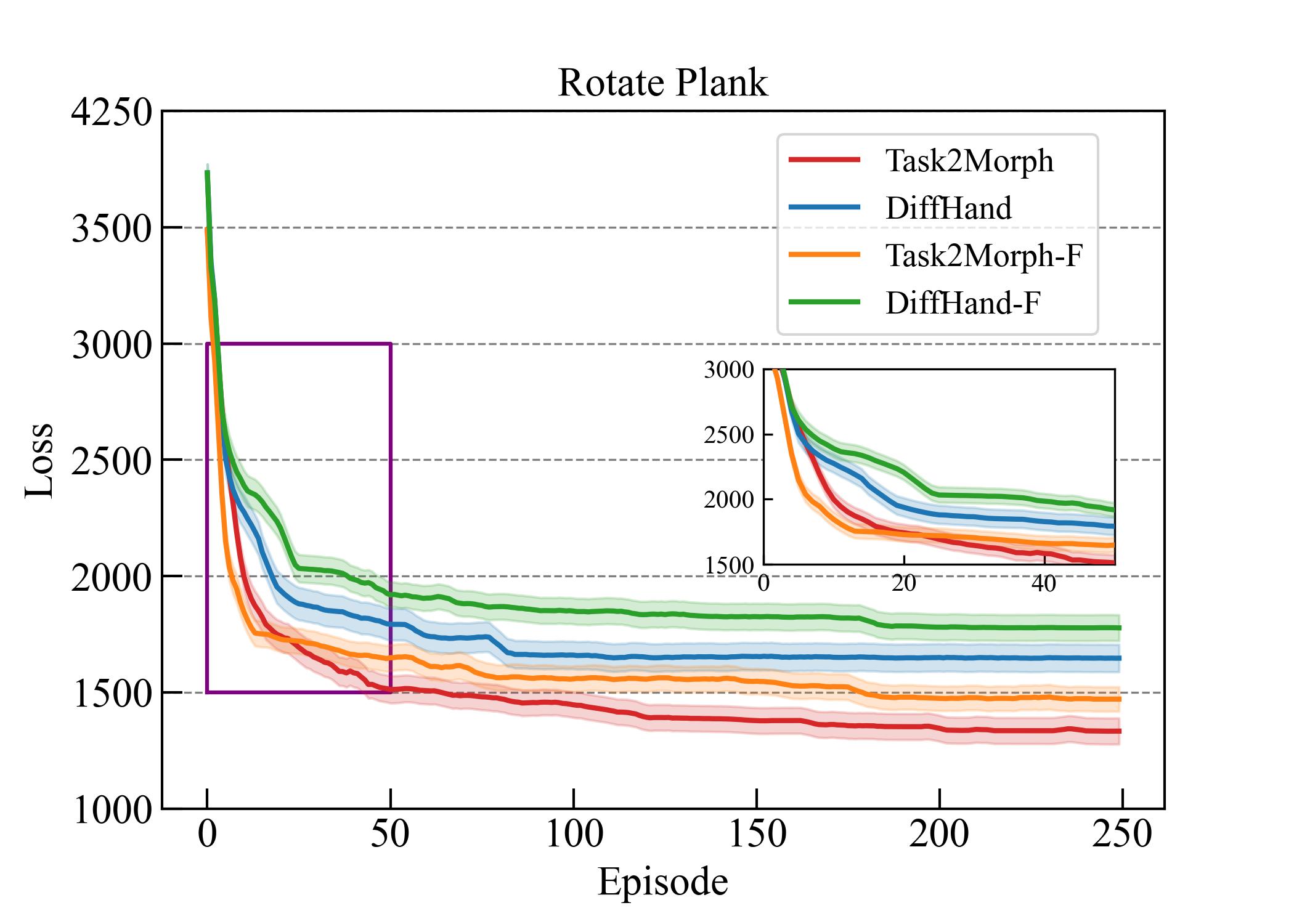} \\
			\end{minipage}
		}
	
		\caption{The average performance of the \emph{Task2Morph}, \emph{Task2Morph-F}, \emph{DiffHand} and \emph{DiffHand-F} in three scenarios. Figures report the mean and standard deviation of the loss for 20 randomly selected tasks in every scenario. Some areas in the subfigures are scaled to clarify the convergence speed of the algorithms. The horizontal axis of each plot stands for the number of simulation episodes, while the vertical is the task-specific objective loss $ \mathcal{L} $. We smooth out the curves with a window size of 5.}
		\label{loss}
	\end{figure*}

	\begin{figure*}[t]
	\centering
	\includegraphics[scale=0.217]{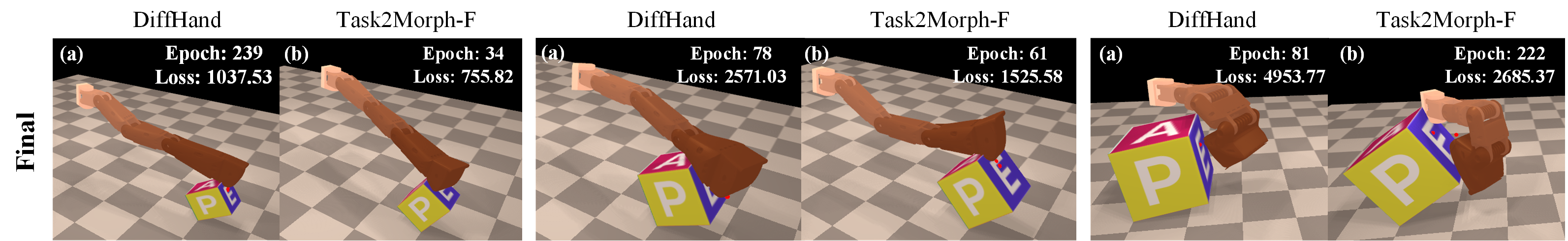}
	\caption{Experimental results of \emph{DiffHand} and  \emph{Task2Morph-F} under three different tasks in \emph{Flip Box} scenario. Each task comprises two subfigures, denoted as (a) and (b), illustrating the optimized final morphology for \emph{DiffHand} and \emph{Task2Morph-F}, respectively.}
	\label{show2}
	\end{figure*}

	\begin{figure*}[t]
		\centering
		\subfigure
		{
			\begin{minipage}[b]{.31\linewidth}
				\centering
				\includegraphics[scale=0.35]{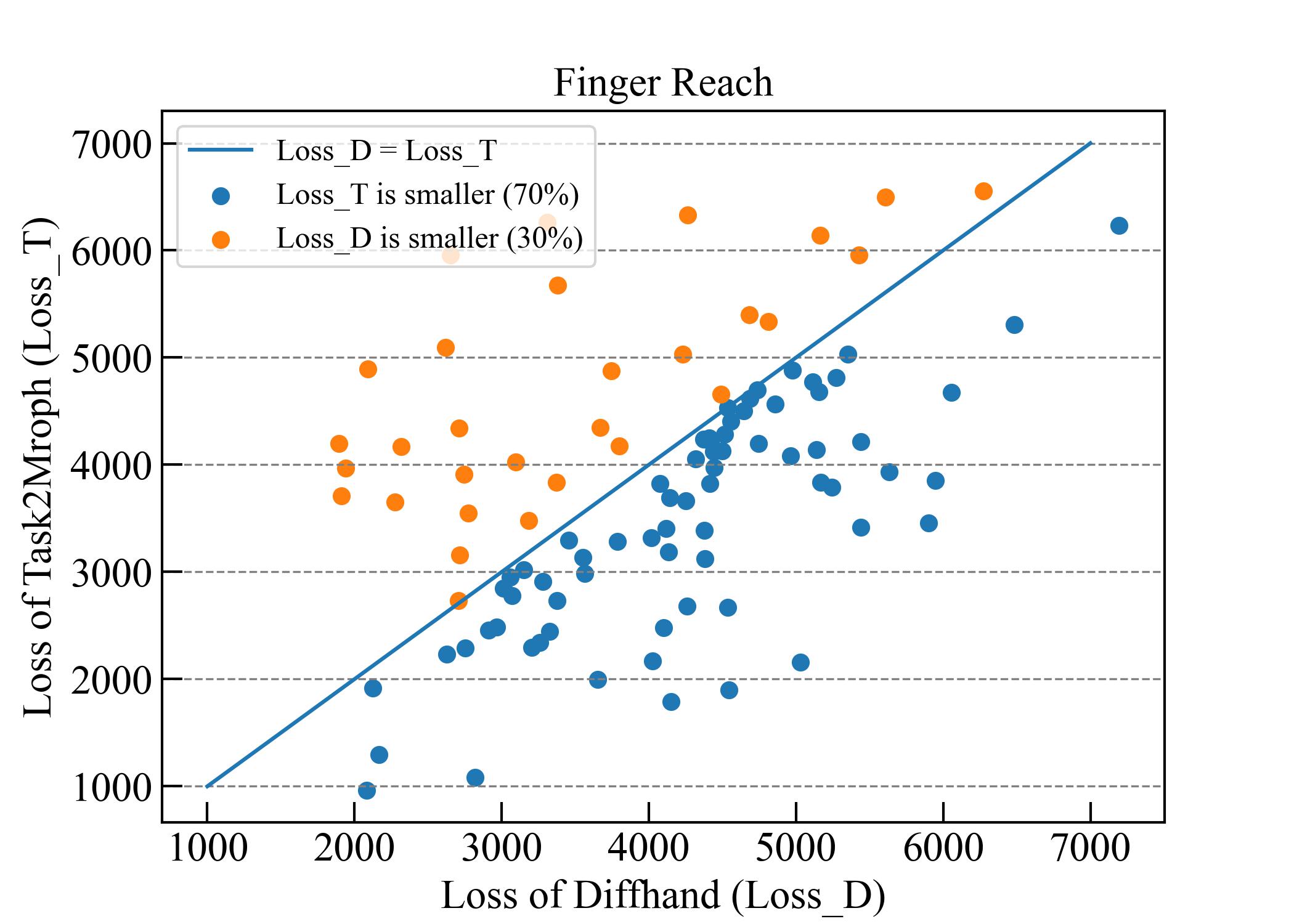} \\
			\end{minipage}
		}
		\subfigure
		{
			\begin{minipage}[b]{.31\linewidth}
				\centering
				\includegraphics[scale=0.35]{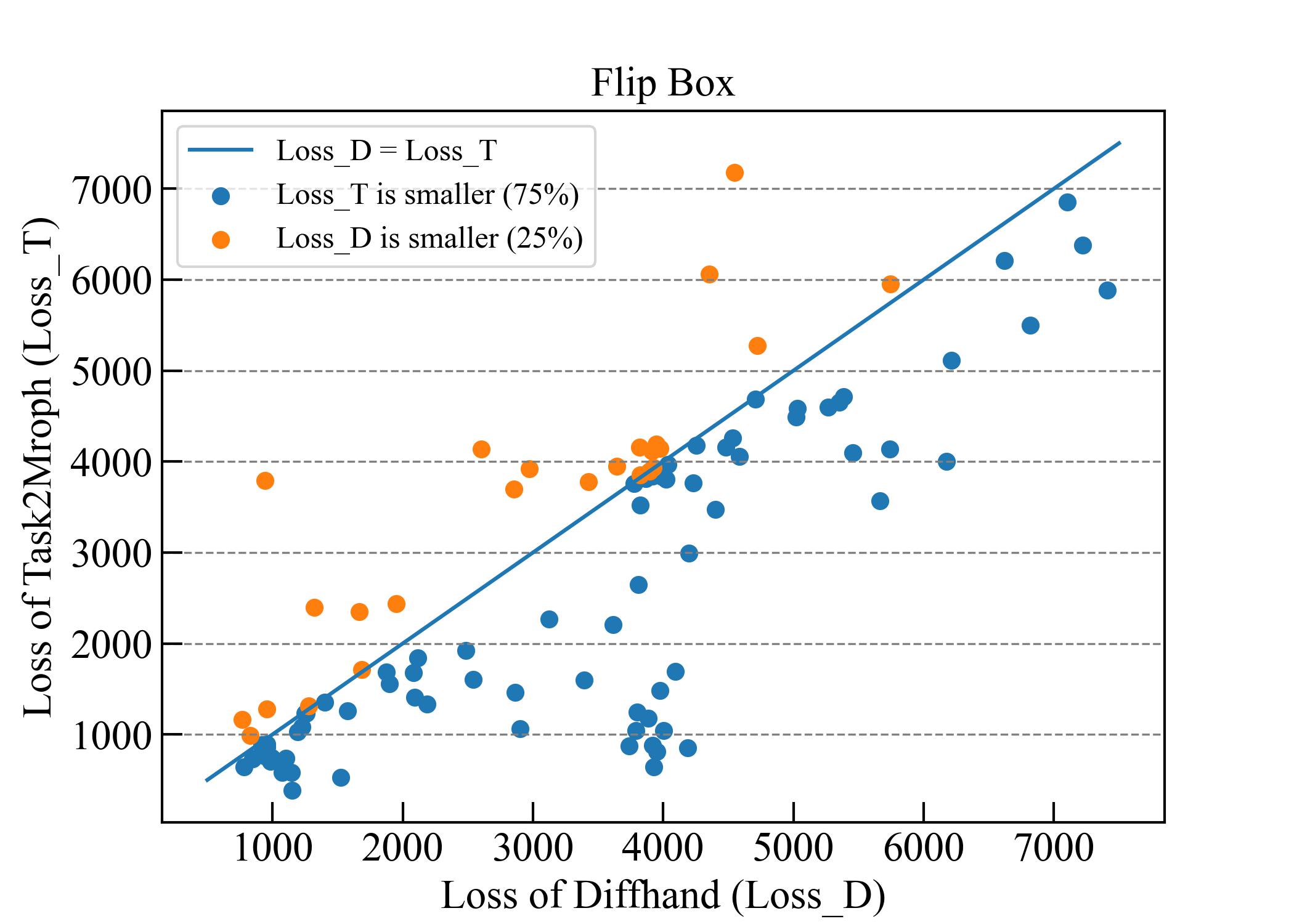} \\
			\end{minipage}
		}
		\subfigure
		{
			\begin{minipage}[b]{.31\linewidth}
				\centering
				\includegraphics[scale=0.35]{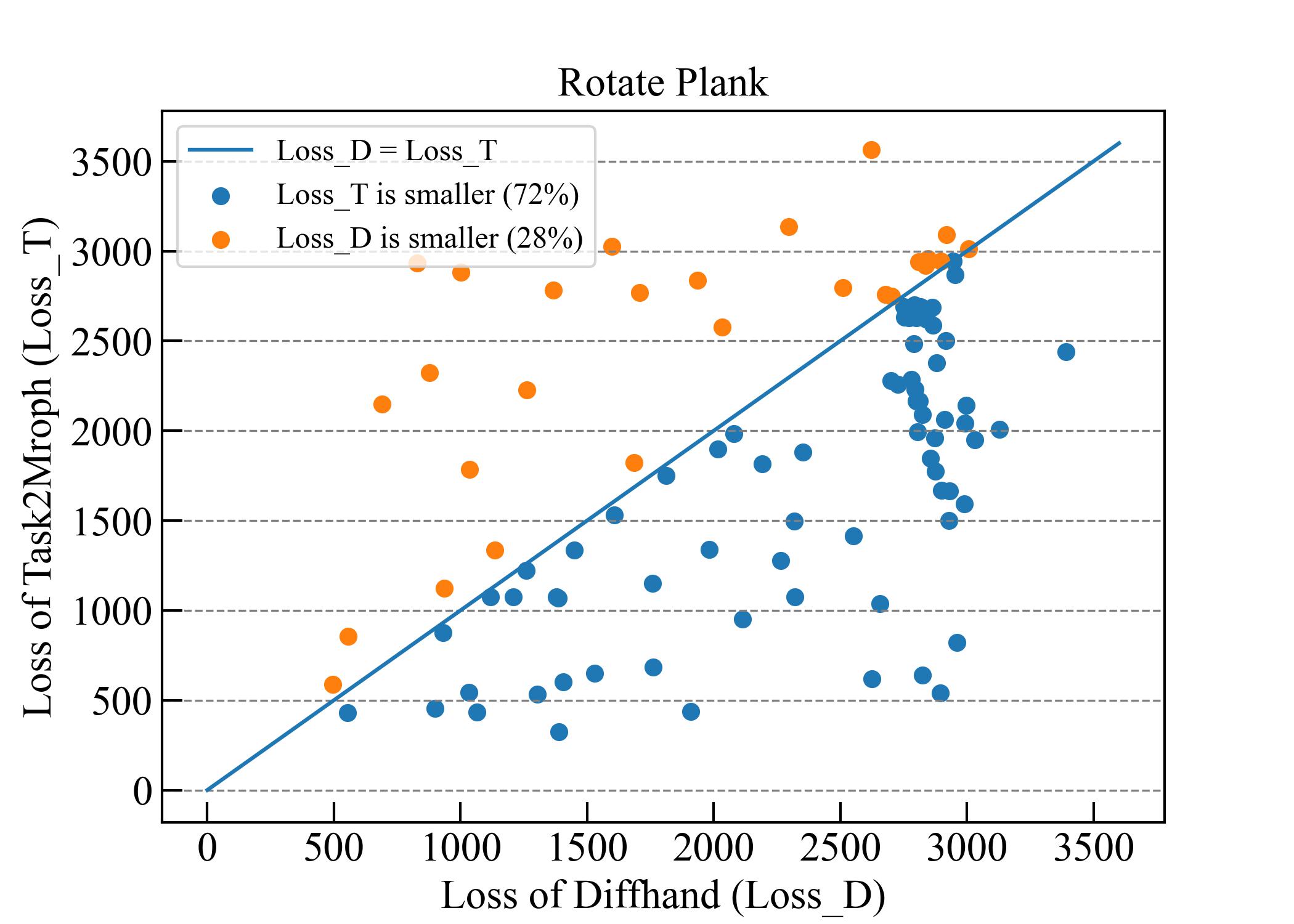} \\
			\end{minipage}
		}
		\caption{Comparison of objective loss of initial morphologies generated by \emph{Task2Morph} and \emph{DiffHand}. Randomly select 100 tasks in  \emph{Finger Reach}, \emph{Flip Box}, and \emph{Rotate Plank} scenarios, respectively, and each task gets the point with the loss of \emph{DiffHand} on the horizontal axis and the loss of \emph{Task2Morph} on the vertical axis when the algorithm converges. 70\%, 75\% and 72\% tasks of \emph{Task2Morph} achieve better performance, respectively, and the average loss value is reduced by about 194.05, 536.01 and 311.17.} 
		\label{scatter}
		
	\end{figure*}

	\section{EXPERIMENTS}
	In this section, we conduct simulation experiments in three scenarios inspired by Xu et al. \cite{xu2021end}: \emph{Finger Reach}, \emph{Flip Box} and \emph{Rotate Plank}. Firstly, we compare the performance of \emph{Task2Morph} and \emph{DiffHand} to demonstrate the improvement of our method in terms of algorithm convergence speed and robot performance. Secondly, we conduct experiments on a diverse set of tasks to showcase the robust adaptability and flexibility of the task-inspired initial morphologies. Lastly, we provide specific examples of robot design results from both frameworks in three scenarios. The code and data links are available at \url{https://github.com/HPCL-EI/Task2Morph.git}.
	



		\begin{figure*}[t]
		\centering
		\includegraphics[scale=0.218]{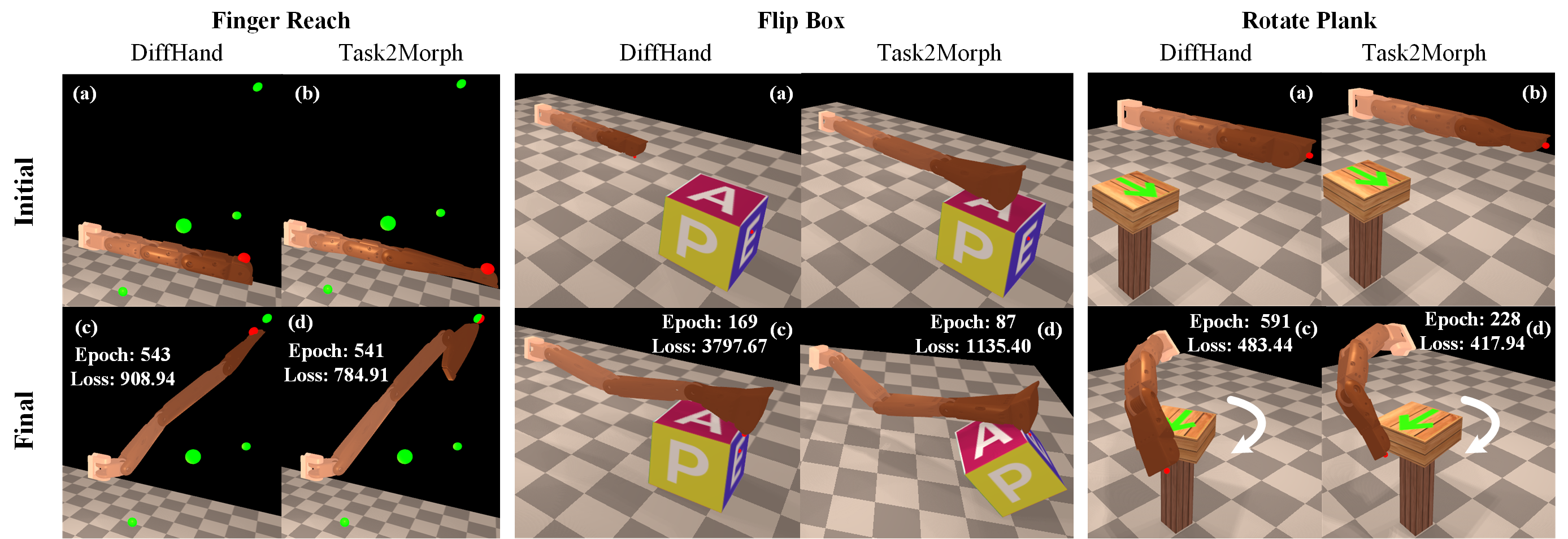}
		\caption{Experimental results of \emph{Task2Morph} and \emph{DiffHand} on one task in each of the three scenarios. Each task has four subfigures labeled (a),(b),(c) and (d), which respectively show the initial morphology of \emph{DiffHand}, the initial morphology of \emph{Task2Morph}, the optimized final morphology of \emph{DiffHand}, and the optimized final morphology of \emph{Task2Morph}. Subfigures (c) and (d) record the number of episodes and the final loss when the algorithm converges.}
		\label{show}
	\end{figure*}
	
	%
	
	\subsection{Experimental Setup}
	
	
	

	\textbf{Tasks and Task Abstraction.} We present the task objectives for the three scenarios in Table \ref{objectives}, where the task-specific objective loss $\mathcal{L}$ is inversely correlated to task completion, as explained in Xu et al. \cite{xu2021end}. To abstract task-relevant features to task completion, we manually selected features such as object location and size across different tasks, as summarized in Table \ref{objectives}. We adjust additional parameters in the simulation environment to enhance task complexity and diversity. For example, in the \emph{Flip Box} scenario, we expand finger control to include two-dimensional free movement instead of limited planar up-down motion. Unlike the fixed task features in \emph{Diffhand}, the task features in this study are variable, with the range of selected object feature variations for the three scenarios presented in Table \ref{objectives}. For further details, please refer to the open-source code of our project.

	
	
	
	
	\textbf{Regression Model Construction and Training.} 
	We use a fully connected neural network based on PyTorch to establish a mapping between the task features and optimal morphology parameters. The network consists of an input layer, two hidden layers with 32 nodes each, and an output layer. The input is the extracted task features, and the output is nine morphology parameters. We employ stochastic gradient descent (SGD) with an experience buffer size of 100 and a first-in-first-out updating strategy. We set the mini-batch size to 10 and the weight coefficient $\alpha$ to 0.3. 
	
	 
	The experiments are performed on a workstation equipped with an AMD Ryzen 9 5900X 12-Core Processor CPU, one NVIDIA GeForce RTX 3090 GPU, and 128GB of memory. We focus on the multi-task adaptability within each individual scenario, rather than considering the inter-scenario adaptability. The average time for collecting and testing one task in the \emph{Finger Flip} and \emph{Rotate Plank} scenarios is approximately one minute, with a total training time of three hours. The \emph{Finger Reach} scenario takes approximately seven times longer than the other two scenarios.
	
	
	\textbf{Baselines.}
	We evaluate and compare the following four baseline algorithms.
	\begin{itemize}
	\item \emph{Task2Morph}: Our task-inspired morphology framework. Furthermore, the regression model has been trained online through joint optimization for 150 random tasks. The robot is capable of generating the corresponding initial morphology based on the features of a new task.
	\item \emph{Task2Morph-F}: Unlike the joint optimization of \emph{Task2Morph}, the initial morphology generated in this method is frozen and no fine-tuning is performed. Only the control sequence can be optimized.
	\item \emph{DiffHand}: The end-to-end differentiable morphology and controller joint optimization by Xu et al. \cite{xu2021end}. For each new task, the initial morphology is the same and no task-inspired generation is involved. The configuration of the initial morphologies follows the approach described in \cite{xu2021end}.
	\item \emph{DiffHand-F}: Unlike \emph{DiffHand}, the initial morphology is fixed and frozen during optimization. Only the control sequences can be optimized.
	\end{itemize}


	\subsection{Performance Evaluation}
	

	
	We randomly sample 20 tasks to evaluate four baseline algorithms. The results depicted in Figure \ref{loss} demonstrate that our method can discover more suitable morphologies and control sequences, and achieve faster convergence of the loss function, particularly in the early stages of the simulation. These benefits are consistent with the embodied intelligence hypothesis, which states that agents with appropriate morphologies can rapidly acquire intelligent behaviors \cite{gupta2021embodied}. Furthermore, our experiments reveal that \emph{Task2Morph-F}, focusing solely on optimizing the controller, outperforms Diffhand in specific tasks. This is particularly evident in the \emph{Flip Box} scenario, where \emph{Task2Morph-F} achieves comparable performance to \emph{Task2Morph} without the need for further fine-tuning. 
	
	
	To provide additional insights, we illustrate three tasks within the \emph{Flip Box} scenario in Figure \ref{show2}. Notably, the final morphology of \emph{Task2Morph-F} remains the same as its initial morphology. This morphology enables the robot to flip the box effectively, outperforming  \emph{Diffhand} in task completion and convergence speed. These findings suggest that by employing a well-trained task-inspired morphology generation model, the additional optimization loop for morphology may be obviated, leading to exponential reductions in the complexity of the two-loop robot design problem.

	\subsection{Task Adaptability and Fine-tuning} 
	
	
	
	We compare our initial morphology with the fixed initial morphology used by \emph{DiffHand}. In the case of optimizing only the controller, the robot attempts 100 random tasks in each scenario, as shown in Figure \ref{scatter}. The results demonstrate that our initial morphologies are more adaptive than fixed morphologies and are naturally better suited for different tasks. Quantitatively, our robots outperform those with fixed initial morphologies on over 70\% of the tasks.

	Although our initial morphology already exhibits task adaptability, fine-tuning can yield better results in some tasks. We select one task in each scenario and visualize the initial and final morphologies of \emph{DiffHand} and \emph{Task2Morph} in these tasks, as shown in Figure \ref{show}. In the \emph{Finger Reach} scenario, the superior initial morphology motivates the robot's end to optimize a complex morphology, such as a fishtail. In the \emph{Flip Box} scenario, our robot is able to flip the box powerfully, whereas \emph{DiffHand} is unable to do so due to its thin joints. In the \emph{Rotate Plank} scenario, our robot is able to rotate the plank 90 degrees with its short end, whereas \emph{DiffHand} requires extensive iterations to achieve the same outcome.

	\section{CONCLUSIONS}
	
	Intelligent and efficient morphology adaptation methods will drive the future of embodied intelligence and robot design. In this work, we propose \emph{Task2Morph}, a novel differentiable task-inspired framework for contact-aware robot design. By abstracting performance-related task features and constructing a task-to-morphology mapping, our framework allows robots to generate appropriate morphologies for different tasks. While we only integrate our task-inspired morphology generation into Xu et al.'s \cite{xu2021end} joint optimization method in this paper, \emph{Task2Morph} can be easily extended or applied to other robot co-optimization methods. It is also particularly suitable for morphology adaptation scenarios where task features vary. Therefore, when combined with intelligent perception algorithms, it has great potential to enable robot morphology adaptation across multiple scenes and tasks. Furthermore, our method can be transferable to real-world scenarios using deformable physical robots.
	

	\section*{ACKNOWLEDGMENT}
	
	This work was supported by the National Natural Science Foundation of China (Grant Nos. 62106278, 91948303-1, 611803375, 12002380,  62101575), the National Key R\&D Program of China (Grant No. 2021ZD0140301).

	\addtolength{\textheight}{-12cm} 
	\bibliographystyle{unsrt}
	\bibliography{hyperref}
	
\end{document}